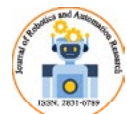

# Diversity of Thought Elicits Stronger Reasoning Capabilities in Multi-Agent Debate Frameworks

**Mahmood Hegazy***

*University of Montreal, Mila - Quebec AI Institute, Canada*

*****Corresponding Author**
Mahmood Hegazy, University of Montreal Mila - Quebec AI Institute, Canada.

## Abstract
Large language models (LLMs) excel in natural language generation but often confidently produce incorrect responses, especially in tasks like mathematical reasoning. Chain-of-thought prompting, self-verification, and multi-agent debate are among the strategies proposed to improve the reasoning and factual accuracy of LLMs. Building on Du et al.'s multi-agent debate framework, we find that multi-agent debate helps at any model scale, and that diversity of thought elicits stronger reasoning in debating LLMs. Across various model sizes, performance on mathematical reasoning tasks benefits most when diverse trained models are used. Remarkably, after 4 rounds of debate, a diverse set of medium-capacity models (Gemini-Pro, Mixtral 7B×8, and PaLM 2-M) outperforms GPT-4 on the GSM-8K benchmark, scoring 91% accuracy. By comparison, when 3 instances of Gemini-Pro are used, performance only reaches 82%. Finally, this diverse set of medium-capacity models sets a new state-of-the-art performance on the ASDiv benchmark (94%). These results underscore the idea that the future of AI is agentic, with diverse cooperating agents yielding emergent capabilities beyond even the most powerful individual models.

## 1. Introduction

In the dynamic realm of artificial intelligence, enhancing the reasoning abilities and factual accuracy of large language models (LLMs) stands as a pivotal challenge. Central to this pursuit is the exploration of innovative methodologies that address existing shortcomings and chart new pathways for advancement. Our research aims to fortify the foundations of LLMs through the lens of multi-agent debate. The key motivation behind this project is solving the issue of "hallucination" within language models, where plausible yet erroneous information is generated, undermining their reliability and trustworthiness. Inspired by the collaborative nature of human intellectual discourse, the methodology of multi-agent debate emerges as a promising solution to this problem. By harnessing the collective insights of multiple AI agents engaged in structured debate, we seek to not only mitigate hallucinations but also elevate the precision and reliability of LLM responses.

A glance at the landscape of current research reveals a series of endeavors aimed at fortifying the reasoning capabilities of LLMs. While recent iterations of LLMs, such as GPT-4, represent significant strides forward, concerns persist regarding their reasoning capabilities. In response to these limitations, advocates the transformative potential of multi-agent debate. Furthermore, advancements in agentic approaches such as MetaGPT, as proposed by, and Agentverse, as proposed by offer diverse perspectives, delving into collaborative problem-solving and simulation of human behavior, thus broadening the horizons of LLM research [1-3].

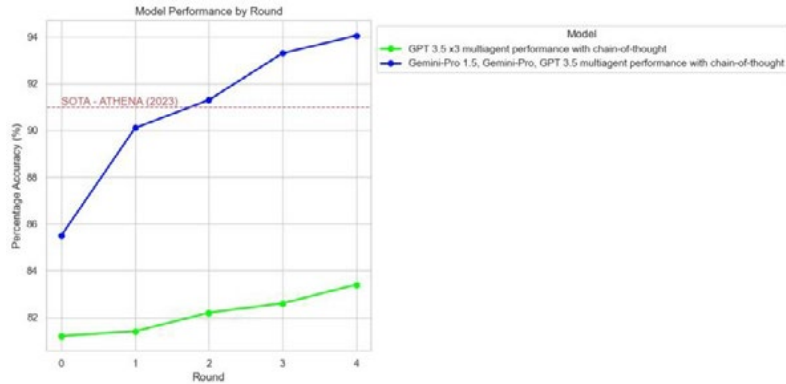

**Figure 1:** Diverse Model Debate Performance Across 4 rounds on the ASDiv Benchmark

Our work aims to mitigate serious LLM deficiencies in reasoning by offering a nuanced analysis of existing multi-agent debate frameworks and their effectiveness in enhancing reasoning in LLMs. Building upon the insight from previous studies, our approach emphasizes the importance of diversity in models and debate strategies (diversity of thought). By synthesizing findings from diverse benchmarks and methodologies, we provide a comprehensive understanding of the strengths and limitations of current frameworks.

To empirically validate the effectiveness of our approach, we performed comprehensive experiments utilizing both diverse and homogeneous sets of language models with varying capacities. These experiments were conducted on multiple mathematical reasoning benchmarks, including the challenging GSM-8K dataset and the recently introduced ASDiv benchmark, which assess the models' ability to generate accurate and well-reasoned solutions to complex problems. Our results demonstrate that leveraging diversity of thought in multi-agent debate significantly enhances the reasoning capabilities of LLMs, outperforming even state-of-the-art models like GPT-4. These findings underscore the potential of diverse, cooperating agents in achieving emergent capabilities beyond individual models.

## 2. Related Work

The landscape of large language models (LLMs) has witnessed remarkable advancements in recent years, exemplified by innovations such as GPT-4, Llama , and PaLM [4-6]. However, despite these breakthroughs, a critical examination reveals significant concerns regarding the reasoning capabilities of LLMs, as highlighted by [7]. This recognition has motivated a series of research efforts aimed at enhancing the reasoning and problem-solving abilities of LLMs through various methodologies.

Approaches like chain-of-thought prompting, self-verification, and multi-agent debate, as introduced by, and respectively, have been proposed to address this challenge [1,8]. The multi-agent debate approach, inspired by Minsky [1988]'s "Society of Mind" theory, posits that diverse agents approaching a problem with different methods, purposes, knowledge representations, and result-production techniques can enhance factual accuracy through debate and communication. Similarly, introduced MetaGPT, a meta-programming framework designed to tackle logic inconsistencies and hallucination by incorporating Standardized Operating Procedures (SOPs) and structured communication within LLM-based multi-agent systems [2].

In parallel, efforts have been directed towards enhancing the generative capabilities of LLMs to simulate believable human behavior, as studied by [9-11]. These endeavors, while successful in creating realistic simulations, have also prompted further exploration into refining retrieval modules and real-time interactivity to mitigate instances of hallucination. Furthermore, frameworks such as Agentverse, as proposed by prioritize collaborative problem-solving among autonomous agents, emulating human group dynamics to achieve superior performance across diverse tasks [3]. This emphasis on collaborative reasoning sets a precedent for future developments aimed at refining the intricacies of LLMs and advancing their capabilities in various domains. Collectively, these studies underscore the imperative to address the reasoning deficits of LLMs while also exploring avenues for enhancing their generative capabilities and facilitating collaborative problem-solving. Our work builds upon these foundations, leveraging the power of multi-agent debate and diversity of thought to push the boundaries of LLM reasoning and pave the way for more reliable and capable language models.

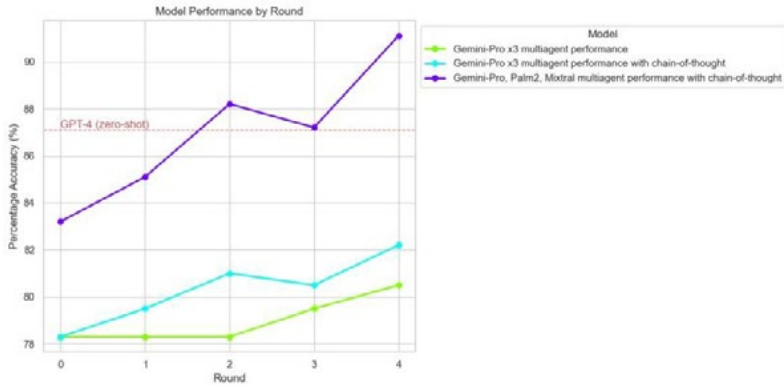

**Figure 2:** Diverse Model Debate Performance Across 4 rounds on the GSM-8K Benchmark

## 3. Methodology

Our multi-agent debate framework for enhancing the mathematical reasoning capabilities of LLMs broadened the scope of the framework, as introduced by to ensure it was compatible with diverse models architectures. As illustrated in Figure 3, it consists of the following key components:

• Question Encoding: The mathematical question or problem is provided as input to the system. This question serves as the starting point for the debate among the participating models.

• Debating Models: Three diverse language models - Model 1, Model 2, and Model 3 - are employed as the debating agents. These models can be chosen to have different architectures. We utilized this architecture to run experiments with and without diverse models.

• Debate Rounds: The debating models engage in multiple rounds of debate, where each model generates a response to the question based on its own reasoning capabilities. In each round, the models take turns providing their responses.

• Response Summarization: After each round, the responses from all three debating models are passed through a fourth model - Model 4 (Response Summarizer). This model's role is to analyze and summarize the key arguments, reasoning steps, and conclusions presented by the debating models. The summarized response captures the most salient and convincing points from the debate round. Our model of choice for response summarization was Gemini-Pro for all experiments.

• Iterative Refinement: The summarized response from Model 4 is then fed back as input to the debating models for the next round of debate. This iterative process allows the models to build upon each other's reasoning, identify and correct errors, and refine their arguments based on the collective insights generated in the previous rounds.

• Final Summary: After a predefined number of debate rounds (n), the final summarized response from the summarizer model (Model 4) is considered as the output of the multi-agent debate framework. This final summary represents the consolidated reasoning and conclusion arrived at through the iterative debate process. Here we can extract what the mode of the answers of the 3 models was.

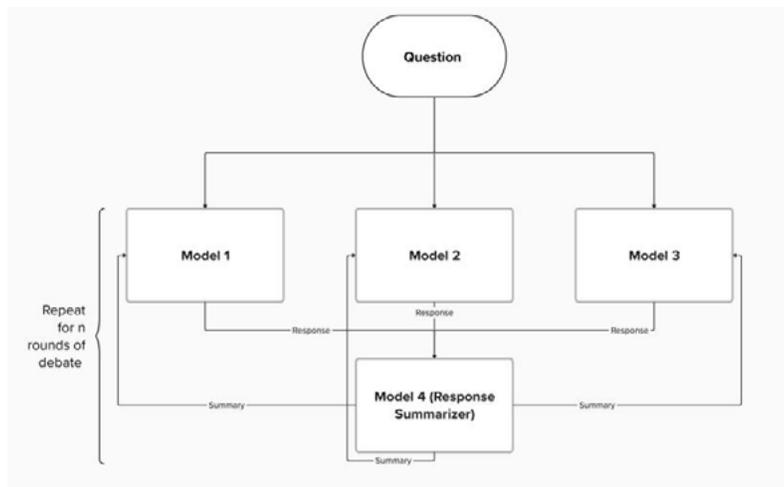

**Figure 3:** Multi Agent Debate Framework Architecture

The debate framework leverages the diversity of the participating models to explore different reasoning paths, challenge assumptions, and arrive at more robust and accurate conclusions. By encouraging the models to critically examine and build upon each other's arguments, the framework aims to mitigate the limitations of individual models and enhance the overall mathematical reasoning capabilities of the system.

### 3.1. Datasets
To empirically evaluate the effectiveness of our multi-agent debate framework, we employ a diverse set of benchmarks that assess various aspects of language understanding and mathematical reasoning.
• GSM-8K This benchmark comprises 8.5K linguistically diverse grade school math word problems, making it ideal for evaluating multi-step mathematical reasoning [12].
• Academia Sinica Diverse MWP Dataset (ASDiv) ASDiv features 2,306 diverse math word problems covering various language patterns and problem types encountered in elementary school. It includes annotations for problem type and grade level [13].
• MATH This historically challenging dataset provides 12,500 competition mathematics problems, each accompanied by step-by-step solutions. It facilitates the teaching of answer derivations and explanations [14].

Through extensive experiments, we analyze the impact of model diversity, debate round count, and model size on reasoning performance. The results provide insights into the optimal configuration of the multi-agent debate framework for achieving superior mathematical reasoning capabilities compared to individual models.

By leveraging these diverse datasets, we aim to comprehensively assess the effectiveness of our approach in enhancing the reasoning capabilities of LLMs across a wide range of problem types, difficulty levels, and language patterns. This rigorous evaluation enables us to draw meaningful conclusions about the potential of multi-agent debate in advancing the state-of-the-art in language understanding and mathematical reasoning.

### 4. Experiments
To validate the effectiveness of our multi-agent debate framework, we conducted a series of experiments using diverse and homogeneous sets of language models with varying capacities. These experiments were performed on multiple mathematical reasoning benchmarks, as outlined in section 3.1, to assess the models' ability to generate accurate and well-reasoned solutions to complex problems. Our two main goals of this study were as follows:
1. Explore the relationship between model capacity and reasoning performance in the context of multi-agent debate.
2. Investigate the impact of model diversity on the reasoning performance of the multi-agent debate framework.

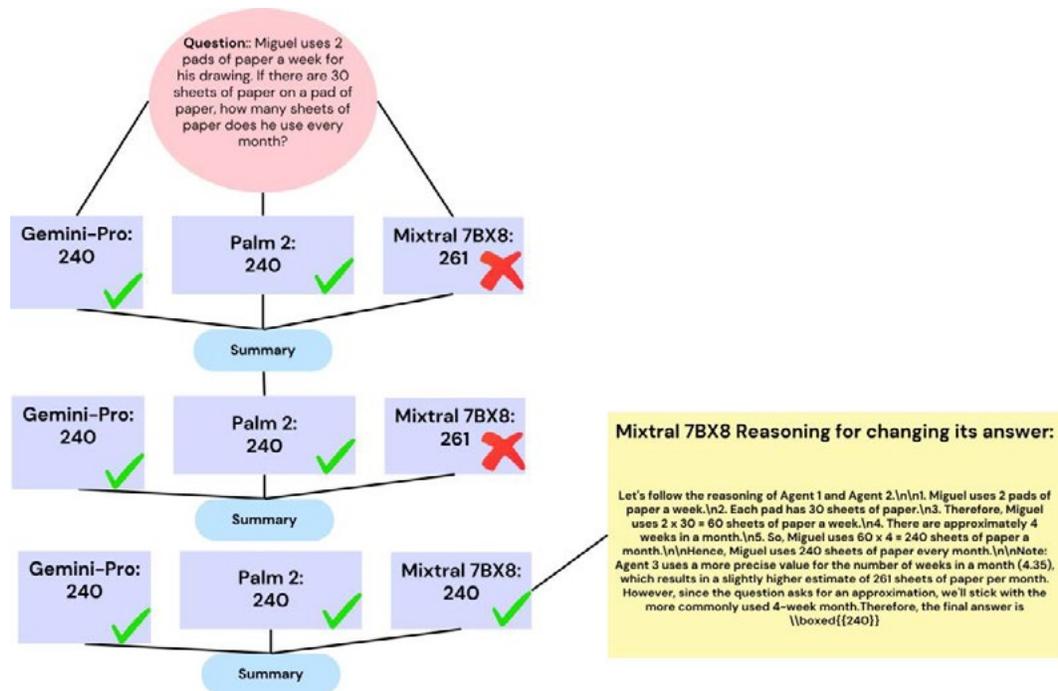

**Figure 4:** Illustration of the Debate Procedure.

**Baseline** To establish a fair and consistent baseline for comparison, we begin by asking each agent to directly generate responses to the given prompts without engaging in debate. This initial response generation serves as round 0 of the debate process and allows us to analyze the performance of the individual models before the collaborative reasoning begins.

By evaluating the models' standalone performance in round 0, we can effectively measure the impact of the subsequent debate rounds on the overall reasoning capabilities of the system. This baseline assessment is crucial for understanding the extent to which the multi-agent debate framework enhances the models' ability to generate accurate and well-reasoned solutions.

To ensure the validity and reliability of our comparisons, we maintain a consistent experimental setup across all evaluations. We use identical starting prompts and language models for both the baseline and the multi-agent debate framework approaches. This consistency eliminates potential confounding factors and enables us to attribute any observed improvements in performance to the effectiveness of the debate process itself.

**Evaluation Methods** To facilitate a comprehensive evaluation of our multi-agent debate framework, we designed a systematic approach to assess the model outputs at each round of the debate. We configured the model prompts to consistently provide a boxed answer at the conclusion of each response, ensuring a standardized format for evaluation.

Following each experiment, we generate a JSON file that encapsulates the model outputs at each round of the debate for every question in the dataset. This structured data allows for a granular analysis of the framework's performance throughout the debate process. Our evaluation script iterates through the debate rounds, assessing the accuracy and quality of the framework's responses at each stage. By comparing the model-generated answers to the ground truth solutions provided in the benchmark datasets, we can quantitatively measure the effectiveness of the multi-agent debate approach in enhancing the reasoning capabilities of LLMs.

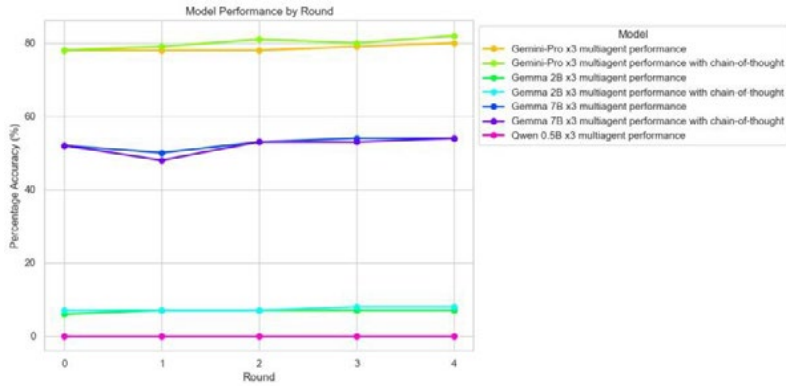

**Figure 5:** Debate Framework Performance across Model Scales 0.75B to 100B+ on GSM8K

Furthermore, we visualize the performance of the framework across rounds, as exemplified in Figure 2. This visual representation provides valuable insights into the progression of reasoning quality as the debate unfolds, allowing us to identify patterns, convergence points, and potential limitations of the approach.

### 4.1. Effect of Model Capacity on Performance
While previous work by explored the effectiveness of multi-agent debate by varying the number of agents and the number of debate rounds, they did not investigate the effect of model scale on the framework's performance [1]. Considering the findings of which demonstrated that chain-of-thought (COT) reasoning is an emergent ability that arises with increased model size, we recognized the importance of conducting a similar experiment for multi-agent debate [8].

To address this gap in the literature, we performed a scaling experiment on the GSM-8K dataset to examine the relationship between model capacity and the performance of the multi-agent debate framework. We evaluated the framework's performance with and without COT reasoning across a range of model sizes, from small-scale models to large-scale ones. The results of our experiment, as presented in Figure 5, revealed a surprising finding.

Contrary to the expectation that multi-agent debate performance would emerge as a direct consequence of increasing model size, we observed similar performance gains across all model scales. This result suggests that the effectiveness of multi-agent debate in enhancing reasoning capabilities is not solely dependent on the model's capacity.

Our findings have significant implications for the development and deployment of multi-agent debate systems. They indicate that even smaller-scale models can benefit from the collaborative reasoning process facilitated by the debate framework, without the need for resource-intensive large-scale models. This insight opens up new possibilities for implementing multi-agent debate in resourceconstrained environments and facilitates the widespread adoption of this approach. Furthermore, our experiment highlights the importance of considering factors beyond model size when designing and optimizing multi-agent debate systems. It suggests that the effectiveness of the debate process may be influenced by other aspects, such as the diversity of the participating models, the structure of the debate, and the quality of the response summarization.

### 4.2. Diversity of Thought
Building upon the previous experiment, we aimed to investigate the impact of model diversity on the performance of the multi-agent debate framework. To introduce diversity, we replicated the study using three models of similar capacity but featuring diverse model families.

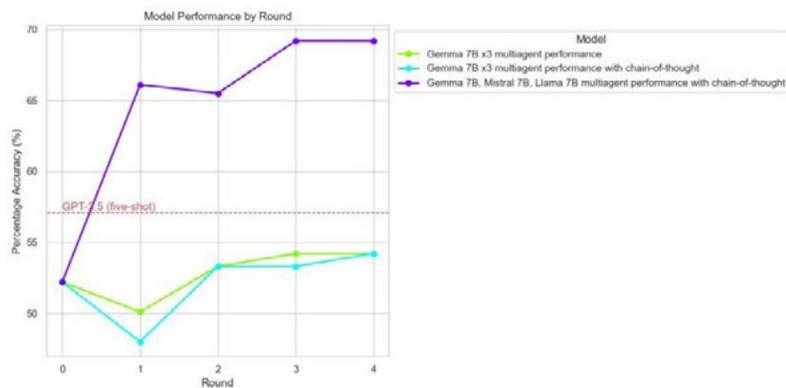

**Figure 6:** 7B Diverse Models Debate Performance Across 4 rounds on GSM8K Dataset

We first evaluated the framework's performance on the GSM8K dataset using Gemini-Pro, PaLM 2-M, and 7x8B (et al. [2024]) [15,16]. When tested individually, these models achieved accuracy rates of 78%, 64%, and 70%, respectively, on the benchmark. Remarkably, the accuracy of the framework improved significantly, rising from 78% to an impressive 91% after 4 rounds of debate, as shown in Figure 2. This substantial improvement outperforms GPT-4 and highlights the power of collaborative reasoning and the synergistic effect of diverse perspectives [17].

To further emphasize the importance of diversity, we compared these results to a homogeneous setup consisting of 3 Gemini Pro models. In the homogeneous case, the performance improved from 78% to 80% without chain-of-thought (COT) reasoning and to 82% with zero-shot COT, as depicted in Figure 2. While still an improvement, the gains in the homogeneous setup were notably lower than those observed in the diverse model configuration.

These findings strongly suggest that the models in the diverse setup greatly benefited from the debate process, leveraging the unique reasoning approaches of their counterparts to refine and enhance their responses. The synergistic effect of combining models with different architectures and capabilities underscores the crucial role of collective insight in boosting overall performance. Our study thus highlights a key insight: within the multi-agent debate framework, diversity is a critical driver of success. By fostering collaboration among models with complementary strengths, the framework enables the emergence of novel reasoning patterns and more accurate solutions to complex problems. Qualitative Result: Figure 4 provides an illustrative example of the dynamics that unfolded during the multi-agent debate experiment. In this particular instance, we observe that Mixtral, one of the participating models, initially maintained its original answer for the first two rounds of the debate.

However, by the third round, Mixtral's stance began to shift as it carefully considered the reasoning put forth by the other two models. This pivotal moment in the debate showcases the power of diverse perspectives in challenging and refining the models' understanding of the problem at hand. Mixtral, before fully adapting its reasoning to align with the collective insights, took a step back to articulate the rationale behind its initial divergent answer. This act of self-explanation not only adds transparency to the debate process but also highlights the model's ability to engage in metacognitive reflection. By acknowledging its initial reasoning and subsequently integrating the persuasive arguments presented by its counterparts, Mixtral demonstrates the capacity for growth and learning within the multi-agent debate framework. This qualitative analysis offers a glimpse into the rich interplay of ideas and the collaborative knowledge construction that emerges when diverse models engage in structured debate. More of this can be found in the appendix.

To further validate the effectiveness of the multi-agent debate framework with diverse models, we conducted additional experiments on the ASDiv and MATH benchmarks. For these experiments, we employed three diverse models: Gemini Flash 1.5, Gemini Pro, and GPT 3.5. On the ASDiv benchmark, the individual performance of these models was already impressive, with Gemini Flash 1.5, Gemini Pro, and GPT 3.5 achieving accuracy rates of 89%, 86%, and 81%, respectively. However, when these models were combined in our multi-agent debate framework, the results were even more remarkable. After 4 rounds of debate, the framework reached an accuracy of 94%, setting a new state-of-the-art performance on the ASDiv benchmark, surpassing the previous record set by as shown in Figure 1 [18]. Similarly, on the challenging MATH benchmark, the individual models achieved accuracy rates of 55%, 32%, and 33% for Gemini Flash 1.5, Gemini Pro, and GPT 3.5, respectively. When engaged in multi-agent debate, the framework's performance significantly improved. By the 4th round of debate, the framework outperformed both GPT-4 and Gemini Ultra by substantial margins of 24% and 14%, respectively, as observed in Figure 8. These results further underscore the power of diverse models in the multi-agent debate framework [19,20]. By leveraging the unique strengths and perspectives of each model, the framework is able to achieve remarkable performance gains, pushing the boundaries of what is possible on these challenging benchmarks.

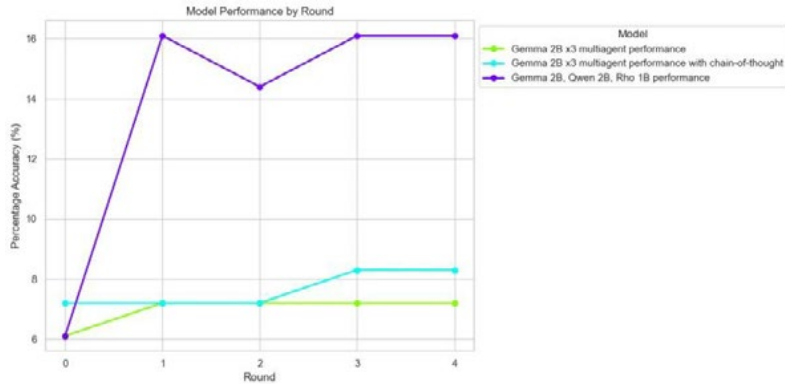

**Figure 7:** 2B Diverse Models Debate Performance Across 4 rounds on GSM8K Dataset

Moreover, to investigate whether diversity of thought is an emergent ability that arises with model scale, we conducted the same diversity experiment on GSM-8K using smaller models. The results were quite notable. Whether using 7B models (Gemma 7B, Mistral 7B, and Llama 2 7B) or 2B models (Gemma 2B, Qwen 2B, and Rho 1B), diversity of thought consistently elicited enhanced reasoning capabilities among these smaller models. As shown in Figures 6 and 7, the 7B model framework achieved a significant performance increase of 17% by the 4th round, while the 2B model framework saw a 10% improvement. These findings challenge our initial hypothesis that the framework's initial competency in the specific benchmark would be crucial for facilitating a productive debate. Instead, our results suggest that the critical requirement for an effective debate is the presence of diverse model architectures of similar capacity, which induces learning and enhances reasoning capabilities.

The consistent performance improvements observed across different model scales highlight the robustness and generalizability of the multi-agent debate framework. The effectiveness of the approach is not limited to large-scale models but extends to smaller models as well, provided that architectural diversity is maintained. This finding has significant implications for the development and deployment of multi-agent debate systems, as it demonstrates the potential for achieving enhanced reasoning capabilities even with resource-constrained models [21,22].

## 5. Conclusion

In this paper, we have presented a comprehensive investigation into the effectiveness of multi-agent debate in enhancing the reasoning capabilities and factual accuracy of large language models (LLMs). By building upon the foundational work of, we have developed an advanced framework that leverages the power of diverse models and iterative refinement to push the boundaries of what is possible in collaborative reasoning and problem-solving [1]. Our experiments on a range of challenging benchmarks, including GSM-8K, ASDiv, and MATH, have consistently demonstrated the remarkable performance gains achieved by the multi-agent debate framework. The diversity of thought inherent in the framework, brought about by the inclusion of models with different architectures and capabilities, has proven to be a critical driver of success. Through the iterative process of debate, the models are able to learn from each other, refine their reasoning, and converge on more accurate and robust solutions.

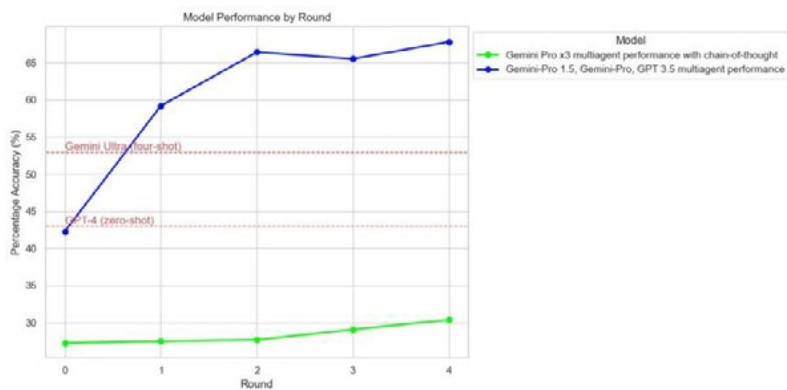

**Figure 8:** Diverse Model Debate Performance Across 4 rounds on the MATH Benchmark

The framework's ability to outperform even state-of-the-art models like GPT-4 and set new records on established benchmarks underscores its potential to revolutionize the field of AI. By harnessing the collective intelligence of diverse models, we have shown that it is possible to achieve emergent capabilities that surpass those of individual models, even the most advanced ones.

Moreover, the consistent effectiveness of the multi-agent debate framework across different model scales and datasets highlights its robustness and generalizability. This versatility opens up exciting possibilities for the application of the framework in various domains, from education and research to industry and beyond. As we look to the future, our findings suggest that the key to unlocking the full potential of AI lies in fostering collaboration and diversity. By embracing the power of multi-agent systems and encouraging the development of models with complementary strengths, we can continue to push the boundaries of what is possible and address increasingly complex challenges.

In conclusion, our work represents a significant step forward in the quest to enhance the reasoning capabilities and factual accuracy of LLMs. The multi-agent debate framework, with its emphasis on diversity and iterative refinement, offers a promising path towards the development of more reliable, trustworthy, and capable AI systems. As we continue to explore and refine this approach, we invite the research community to build upon our findings and join us in shaping the future of collaborative agenitc AI [21].

15. Rohan Anil et al. Palm 2 technical report, 2023.
16. Albert Q. Jiang et al. Mixtral of experts, 2024.
17. Bubeck, S., Chandrasekaran, V., Eldan, R., Gehrke, J., Horvitz, E., Kamar, E., ... & Zhang, Y. (2023). Sparks of artificial general intelligence: Early experiments with gpt-4. *arXiv preprint arXiv:2303.12712.*
18. Kim, J. B., Kim, H., Hahn, J., & Han, Y. S. (2023). ATHENA: Mathematical reasoning with thought expansion. *arXiv preprint arXiv:2311.01036.*
19. Minsky, M. (1988). Society of mind. Simon and Schuster.
20. Weng, Y., Zhu, M., Xia, F., Li, B., He, S., Liu, S., ... & Zhao, J. (2022). Large language models are better reasoners with self-verification. *arXiv preprint arXiv:2212.09561.*
21. Zhang, P., Zeng, G., Wang, T., & Lu, W. (2024). Tinyllama: An open-source small language model. *arXiv preprint arXiv:2401.02385.*